\documentclass[journal]{IEEEtran}
\usepackage{subfigure}

\ifCLASSINFOpdf
\usepackage[pdftex]{graphicx}
\else
\fi
\ifCLASSOPTIONcompsoc
\usepackage[caption=false,font=normalsize,labelfont=sf,textfont=sf]{subfig}
\else
\usepackage[caption=false,font=footnotesize]{subfig}
\fi
\begin{document}
	%
	\title{Extreme Low-Resolution Activity Recognition\\
		with Confident Spatial-Temporal Attention Transfer}
	%
	%
	%
	
	\author{
		Yucai Bai,
		Qin Zou, \IEEEmembership{Senior Member,~IEEE,}
		Xieyuanli Chen, 
		Lingxi Li, \IEEEmembership{Senior Member,~IEEE,} \\
		Zhengming Ding, \IEEEmembership{Member,~IEEE,}
		Long Chen*, \IEEEmembership{Senior Member,~IEEE,}          
		\thanks{Corresponding author: Long Chen.}
		\thanks{
			Y. Bai is with the College of Computer Science, Sichuan University, Chengdu 610012, Sichuan, China.
			(e-mail: 2017223045213@stu.scu.edu.cn)
		}
		\thanks{
			Q. Zou is with the School of Computer Science, Wuhan University, Wuhan 430072, China (e-mail: qzou@whu.edu.cn)
		}
		\thanks{
			X. Chen is with the University of Bonn, Germany (e-mail: xieyuanli.chen@igg.uni-bonn.de)
		}
		\thanks{L. Li is with the Department of Electrical and Computer Engineering, Indiana University-Purdue University Indianapolis, Indianapolis, IN 46202-5132 USA (e-mail: ll7@iupui.edu).
		}
		\thanks{Z. Ding is with the Department of Computer Science, Tulane University, New Orleans, LA 70118, USA (e-mail: zding1@tulane.edu).
		}
		\thanks{L. Chen,  is with the School of Data
			and Computer Science, Sun Yat-Sen University, Guangzhou 518001,
			China (e-mail: chenl46@mail.sysu.edu.cn).
		}
	}
	
	%
	%

	\markboth{Journal of  Image Processing}%
	{}
	%



	\maketitle
	
	\begin{abstract}
		Activity recognition on extreme low-resolution videos, e.g., a resolution of 12$\times$16 pixels, plays a vital role in far-view surveillance and privacy-preserving multimedia analysis. Low-resolution videos only contain limited information. Given the fact that one same activity may be represented by videos in both high resolution (HR) and extreme low resolution (eLR), it is worth studying to utilize the relevant HR data to improve the eLR activity recognition. In this work, we propose a novel Confident Spatial-Temporal Attention Transfer (CSTAT) for eLR activity recognition. CSTAT can acquire information from HR data by reducing the attention differences with a transfer-learning strategy. Besides, the confidence of the supervisory signal is also taken into consideration for a more reliable transferring process. Experimental results demonstrate that, the proposed method can effectively improve the accuracy of eLR activity recognition and achieve an accuracy of 59.41\% on 12$\times$16 videos in HMDB51, a state-of-the-art performance.
	\end{abstract}
	
	
	%
	\IEEEpeerreviewmaketitle

	\section{Introduction}
	\label{intro}

	\begin{figure}[t]
		\centering
		\includegraphics[width=0.48\textwidth]{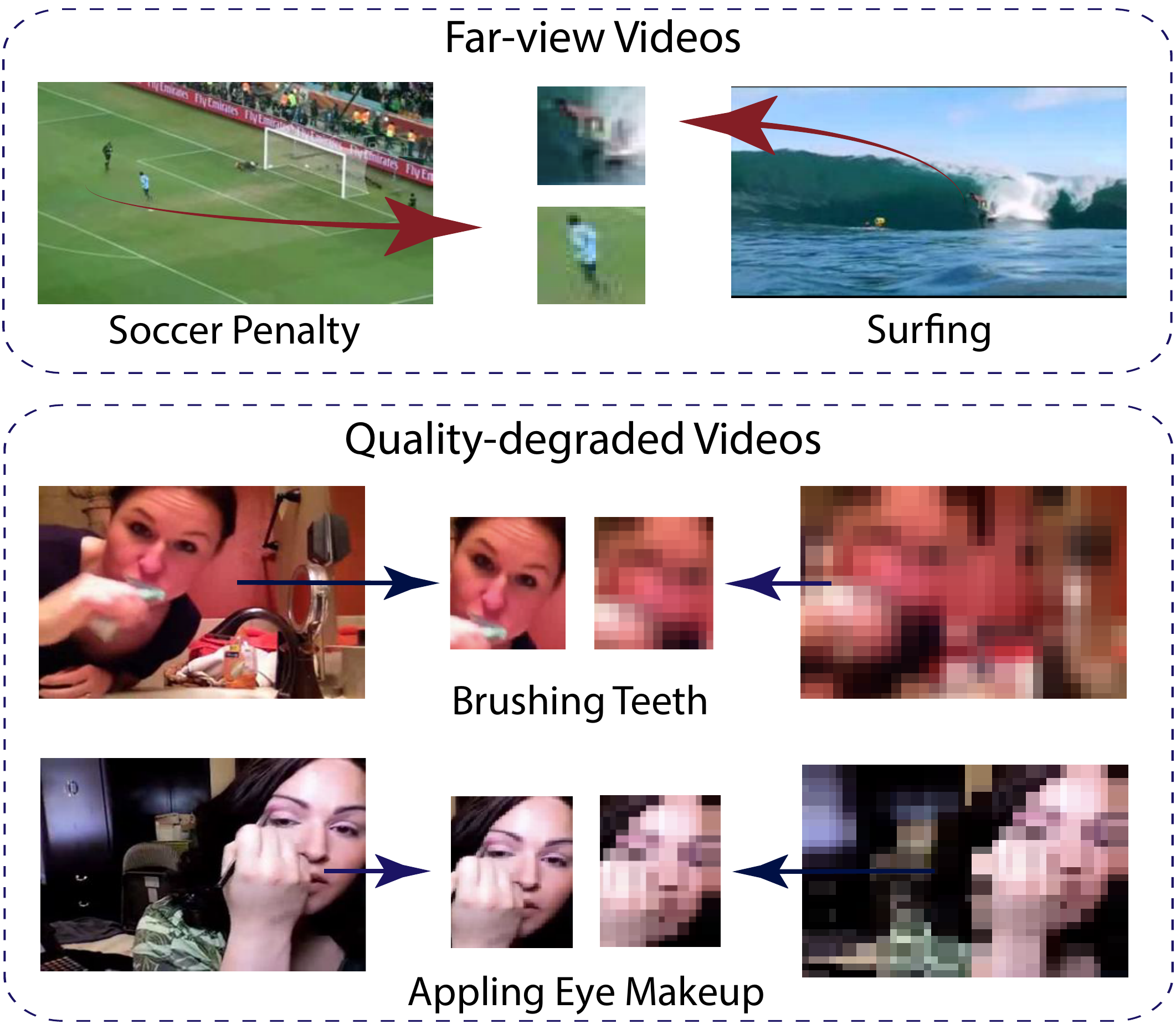}
		\caption{Activities in HR images and eLR images. In far-view videos, activities can only be captured with very few pixels. In quality-degraded videos, activities are represented with blur eLR images on a privacy-preserving purpose.}
		\label{fig:firstsight}
	\end{figure}
	
	Recently, a large number of methods have been proposed for action recognition, and achieved considerable success, e.g., Two-Stream~\cite{simonyan2014two}, C3D~\cite{tran2015learning}, and LRCN~\cite{donahue2015long}, etc. Most of these methods assume that, there is enough information in the region of raw pixels where the activities take place in image sequences. In such cases, action recognition models can well capture and represent the motions of the movement and classify the actions. However, this assumption does not always hold in practice due to the existence of low-resolution and low-quality videos. For instance, in the situation of far-view surveillance, there are often only few pixels for the targets (human in most cases) in the scene. Two examples are shown in the top row in Fig.~\ref{fig:firstsight}. For action recognition in such eLR videos, current popular methods~\cite{simonyan2014two}~\cite{tran2015learning}~\cite{donahue2015long} cannot obtain an acceptable accuracy.
	
	Another situation that can possibly generate eLR videos is in privacy-preserving surveillance or data recording. With the advancement of mobile Internet and smartphones, videos can be easily captured and quickly spread. Meanwhile, average persons would have more concerns about privacy preserving on the web. While videos are captured by ubiquitous cameras and stored in the cloud server, the concern is more urgent. Many technologies have been developed to deal with it, e.g., selective image blurring~\cite{boyle2000effects}, obscuring~\cite{raval2014markit}, etc. A straightforward and effective way to address this problem is to directly collect eLR data during the recording, which can also reduce the video-recording cost and the storage capacity. However, such eLR videos make vision-based tasks more challenging. For example, the bottom of Fig.~\ref{fig:firstsight} shows two images that are degraded for the privacy-preserving purpose. The face in the eLR image is less than 25 pixels, which leads to a great challenge for face recognition.
	
	Benefit from image sequential information, action recognition is still feasible even with eLR videos. Many previous works~\cite{chen2017semi}~\cite{xu2018fully}~\cite{ryoo2018extreme} achieved considerable good results on 12$\times$16 resolution videos, in detection of some specific actions. When the area corresponding to an activity is reduced to few pixels, the part requiring attention will change. The amount of data in eLR images, e.g., 12$\times$16, is about 0.2\% of that in the original 256$\times$340 images. With enough pixels, conventional algorithms can obtain low-level features by capturing geometry and texture details, and a large number of convolution operations and supervision signals from labels can be used to produce high-dimensional features. However, eLR data often do not have enough details for low-level features construction, which brings a bottleneck to the existing algorithms.
	
	How to extract effective information only from few pixels? An effective strategy is to explore the corresponding HR data. We make the assumption that, for the same action, the motion patterns in the eLR videos have some relation with the motion patterns in the corresponding HR videos. The features of an action in the eLR video may be clustered in one space, while features in the HR video may be clustered in another space. However, the distributions of these features in different space may be related using transfer learning. As such, action recognition in eLR videos can benefit from the information of the corresponding HR videos. Most previous researches~\cite{chen2017semi}~\cite{xu2018fully}~\cite{ryoo2018extreme}~\cite{ryoo2015pooled}~\cite{ryoo2017privacy}  adopted this assumption for the higher recognition accuracy in eLR videos.
	
	We also observe that, most of the existing methods for eLR action recognition cannot achieve an adequate accuracy for real-world applications. One reason is that they do not obtain the most useful information from few pixels and do not focus on the most informative area. The inessential areas seldom contain any vital clues for the recognition. In our strategy, these essential details could be obtained from the HR data. We focus on the more informative area, which can help our method fulfill the recognition task. suitable method to describe these areas, which is explored by \cite{jin2019deep}~\cite{zagoruyko2016paying}. For example, the temporal intervals of swinging a golf club and the areas corresponding to the activity are more meaningful than others in determining whether it is an action of playing golf.
	
	Based on the discussion above, we explore the confident spatial-temporal attention transfer model to help recognize actions at eLR videos in this paper. Specifically, we propose a novel knowledge transfer model to utilize more accurate signals from a powerful teacher network. The teacher network uses higher resolution image frames as input, and helps a student network to distinguish which frames are more effective and which parts are more informative for action recognition. 
	Besides, the confidence score in transferring is evaluated to provide extra knowledge, which leads to a high-confidence transferring process. We evaluate the proposed method on well-known datasets and achieve promising results.
	
	The contributions of this work are summarized as follows:
	\begin{itemize}
		\item A simple yet effective algorithm is proposed for measuring the spatial-temporal attention in eLR videos, which is incorporated into the whole knowledge-transfer model for improving the eLR action recognition performance.
		\item To ensure the high quality of the transferring process, confidence is introduced. Three methods of measuring confidence score are discussed. The introduction of estimating confidence would avoid invalid supervision signals caused by the down-sampling process to a large extent.
		\item The proposed method is evaluated on several low-resolution activity benchmarks by comparing with the state-of-the-art methods. An state-of-the-art accuracy of 59.41\% is obtained on 12$\times$16 HMDB51 videos.
	\end{itemize}
	
	This paper is organized with the following structure: In Sec.~\ref{sec:2}, we briefly introduce the three topics of action recognition, teacher-student learning, and eLR vision understanding. In Sec.~\ref{sec:3}, the proposed method is demonstrated in detail. We start from the acquisition of attention, followed by the transferring process, and finally, numerous training strategies are used to make the training more smoothly. In Sec.~\ref{sec:4}, we conducted a wealth of experiments to prove the effectiveness of our proposed method.
	
	\section{Related Work}
	\label{sec:2}
	
	The approach we proposed aims to provide the eLR recognition model with more effective information from the HR data. This section briefly reviews related works from three aspects: activity recognition, teacher-student learning, and extreme low-resolution visual understanding.
	
	\subsection{Activity Recognition}
	
	Recently, activity recognition has undergone tremendous changes, from hand-designed features ~\cite{laptev2003space-time}~\cite{dollar2005behavior}~\cite{sadanand2012action}~\cite{wang2013action}~\cite{veeriah2015differential} to end-to-end deep network methods~\cite{yang2020temporal}~\cite{shao2020intra}~\cite{li2020tea}. This shift occurred due to the introduction of large-scale video datasets~\cite{soomro2012ucf101}~\cite{kuehne2011hmdb} and powerful parallel computing hardware.
	
	Two-Stream network~\cite{simonyan2014two}, as a milestone work in motion recognition algorithms, took the RGB and optical flow for feature extraction, respectively, and then carried out joint estimation, while C3D~\cite{tran2015learning} as a representative work of implicitly extracting temporal information, recognizes continuous frames without processing as input.  Based on these work, 3D convolutions are decomposed into a Pseudo-3D convolutional block as in P3D~\cite{qiu2017learning} or factorized convolutions as in R2Plus1D~\cite{qiu2017learning} or S3D~\cite{xie2017rethinking}, which can speed up the training and inference process of the model without greatly reducing the accuracy. Carreira and Zisserman~\cite{carreira2017quo} proposed to inflate 2D convolutional networks pre-trained on images to 3D for video classification.  ResNeXt~\cite{hara2018can} and MFNet~\cite{chen2018multi} have also been proposed to reduce runtime. RNN as an effective means to exploit temporal clues is also applied in action recognition algorithms, such as LRCN~\cite{donahue2015long}. As an excellent technique for non-structural processing, graph convolution is also widely used in this field by analyzing and recognizing skeleton information more effectively~\cite{liu2020disentangling}~\cite{gao2020drg}.
	
	\subsection{Teacher-Student Learning}
	
	Teacher-Student (TS) learning, as a distillation method, has been widely used. These algorithms could be classified by the number of stages, most of which are divided into two parts. Methods with less or more than two stages have also been proposed. On one hand, TS learning can be a one-stage strategy~\cite{zhang2018deep}~\cite{yang2019snapshot}~\cite{mirzadeh2020improved}. Zhang et al.~\cite{zhang2018deep} proposed that a group of untrained student networks with the same structure can be exploited to learn the target task simultaneously, instead of using the traditional two-stage knowledge extraction strategy. For another, a series of researches~\cite{yang2019snapshot}~\cite{mirzadeh2020improved} focused on TS learning methods in more than two stages. In \cite{mirzadeh2020improved}, multiple teacher assistant networks were used to transfer teachers’ knowledge more easily and effectively. The teacher network first transfers its knowledge to an auxiliary network and then spreads the distilled knowledge to the final student network.
	
	Typically, TS learning is a two-stage method where the teacher network is trained first, then the student network is trained with extra information provided by the teacher network. Hinton et al~\cite{hinton2015distilling} pioneered the concept of knowledge distillation, in which soft targets with more information obtained from the teacher network can help students learn better. Recently, many works have been improved in the way of information dissemination or through the optimization of strict control on the dissemination of information~\cite{zagoruyko2016paying}~\cite{peng2019correlation}~\cite{romero2015fitnets}~\cite{han2020neural}~\cite{tung2019similarity}~\cite{yim2017a}~\cite{park2019relational}. For example, Peng et al~\cite{peng2019correlation} proposed a student network, which not only focuses on imitating the teacher at the instance level but also learns the embedding space, so that the student network could perform better in feature extraction. In addition, there are works studying the role of different teacher networks \cite{ba2014do}~\cite{sau2016deep}~\cite{kang2020towards}. For example, \cite{kang2020towards} used Neural Architecture Search to obtain experience about the architecture from different teacher networks. In addition to image classification tasks, TS learning can also be used in many other different fields, such as face recognition \cite{ge2020efficient}, visual problem solving  \cite{mun2018learning}, video tasks \cite{li2020spatiotemporal}, etc. In summary, teacher networks always have a more complex and deeper network structure to empower the performance of extracting features, while our work enhances the ability of teacher networks by using more informative input.
	
	\subsection{ELR Visual Understanding}
	
	ELR visual tasks are mainly divided into two types, image matching and activity recognition. In terms of image matching, applications with widespread attention are identifying people in the far-view surveillance videos or in the wild. The captured image is always small, and the search pool is actually at different resolutions. One is LR-to-HR, which matches the face of the eLR images with HR database (such as passport photos)~\cite{biswas2010multidimensional}~\cite{shekhar2011synthesis}~\cite{renchuan-xian2012coupled}. \cite{biswas2010multidimensional} embed the eLR images and the HR stored images into a common space, so that their distance is better estimated in this transformed space. In \cite{shekhar2011synthesis}, a generative model for classifying captured images was proposed, which leverages the information available in the HR database. The other is LR-to-LR, where both the captured and the matched are LR facial images~\cite{wang2003face}~\cite{gunturk2003eigenface}~\cite{fookes2012evaluation}~\cite{zou2012very}~\cite{wang2016studying}. \cite{gunturk2003eigenface} transferred the super resolution reconstruction from the pixel domain to the low dimensional face space to solve this problem.
	
	As for eLR activity recognition, the purpose of these works is to understand the far-view surveillance video and protecting privacy. \cite{chen2017semi} proposed a semi-coupled and later-sharing network, which provides the eLR model with more informative clues by sharing part weights.  ~\cite{ryoo2018extreme} used the embedding technique of different downsampling methods, which encodes the mapping strategies in their models. ~\cite{Purwanto_2019_ICCV} used the distillation and exploited the soft targets encoded by teacher network, which could squeeze the analysis results from the teacher network. ~\cite{purwanto2019three} exploited the role of attention as a feature input in eLR visual understanding. However, these methods do not focus on the informative area, which is one of the reasons they can not achieve satisfactory results. 
	
	\section{The Method}
	\label{sec:3}
	
	In this section, we describe the proposed method to recognize multi-class actions from eLR videos in details. We first introduce the acquisition of attention from a single model, which illustrates how to capture the attention distribution from the abundant data. Then, we describe the spatial-temporal attention transfer, where the powerful model imparts more effective attention to the weaker counterpart as a supervision signal. Thirdly, three methods to estimate the confidence score of the transferring are presented. The introduction of transferring confidence could largely reduce the misleading the supervision signals. Finally, the unsupervised training strategies and the dynamically adjusting weights technique are presented, which could solve the over-fitting and help the network learn easier from multiple monitoring signals.
	
	\subsection{Attention Acquisition}
	
\begin{figure}[h]
		\centering
		\includegraphics[width=1.0\linewidth]{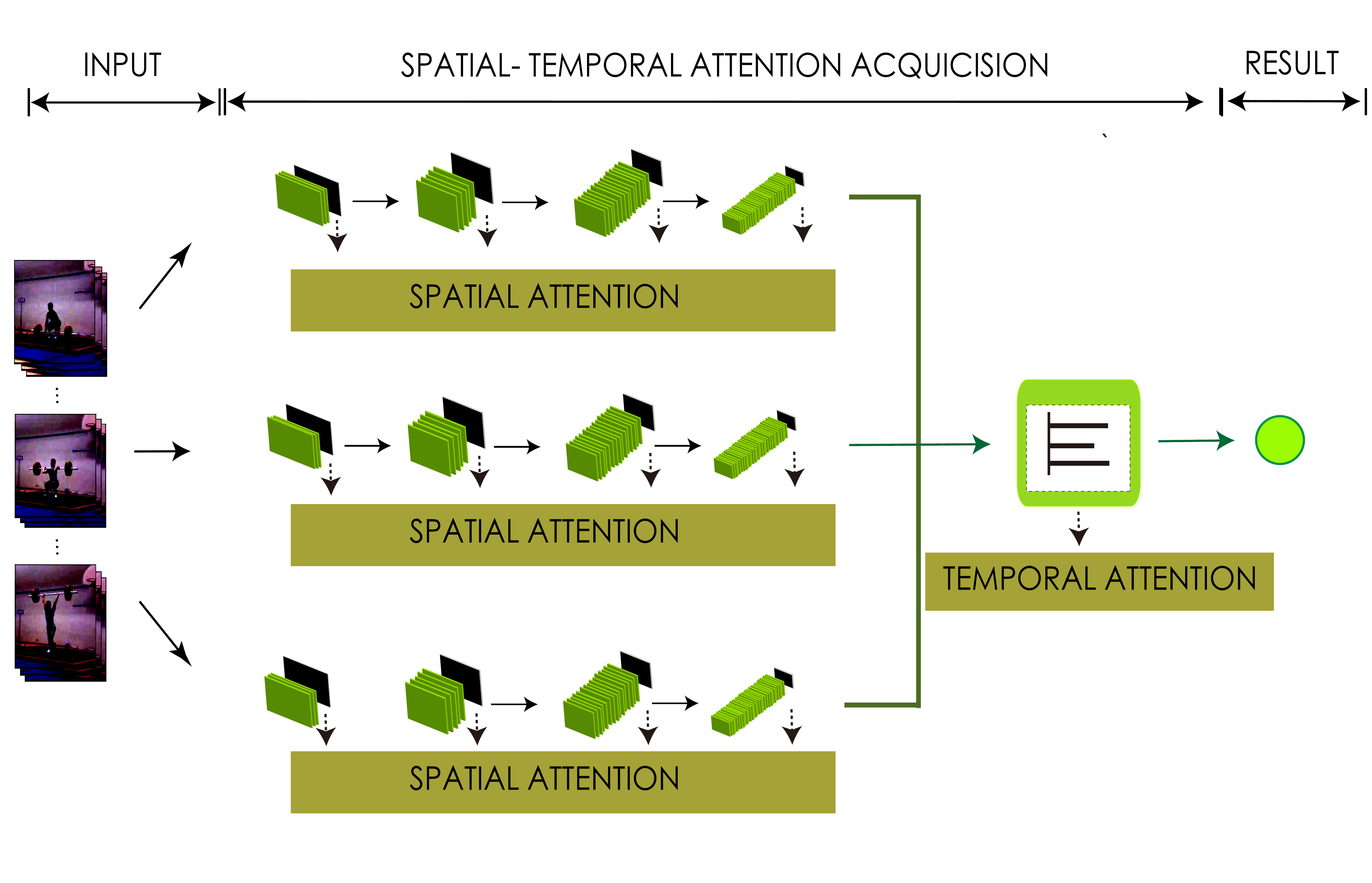}
		\caption{Overview of spatial-temporal attention acquisition. The spatial attention is obtained in feature extraction, and a self-attention module is used to estimate the temporal attention. }
		\label{fig:AA}
	\end{figure}

	The premise of the transferring is to obtain the attention of the network at first. Attention is considered from two aspects: spatial dimension and temporal dimension, as shown in Fig.~\ref{fig:AA}.  Spatial attention refers to the spatial area from which the model classifies the action, might contain the human and corresponding paraphernalia, such as golf club or soccer, while temporal attention means the keyframes containing important action features in the whole videos. 
	
	The proposed method is based on an assumption that the numerical value of the output holds the clues about the model attention and would influence the final classification. In this assumption, the output of middle layers contains the related information, and we want to decode that with our method, which is described as follows.
	
	\subsubsection{Spatial Attention Definition} 
	
\begin{figure}
		\includegraphics[width=0.48\textwidth]{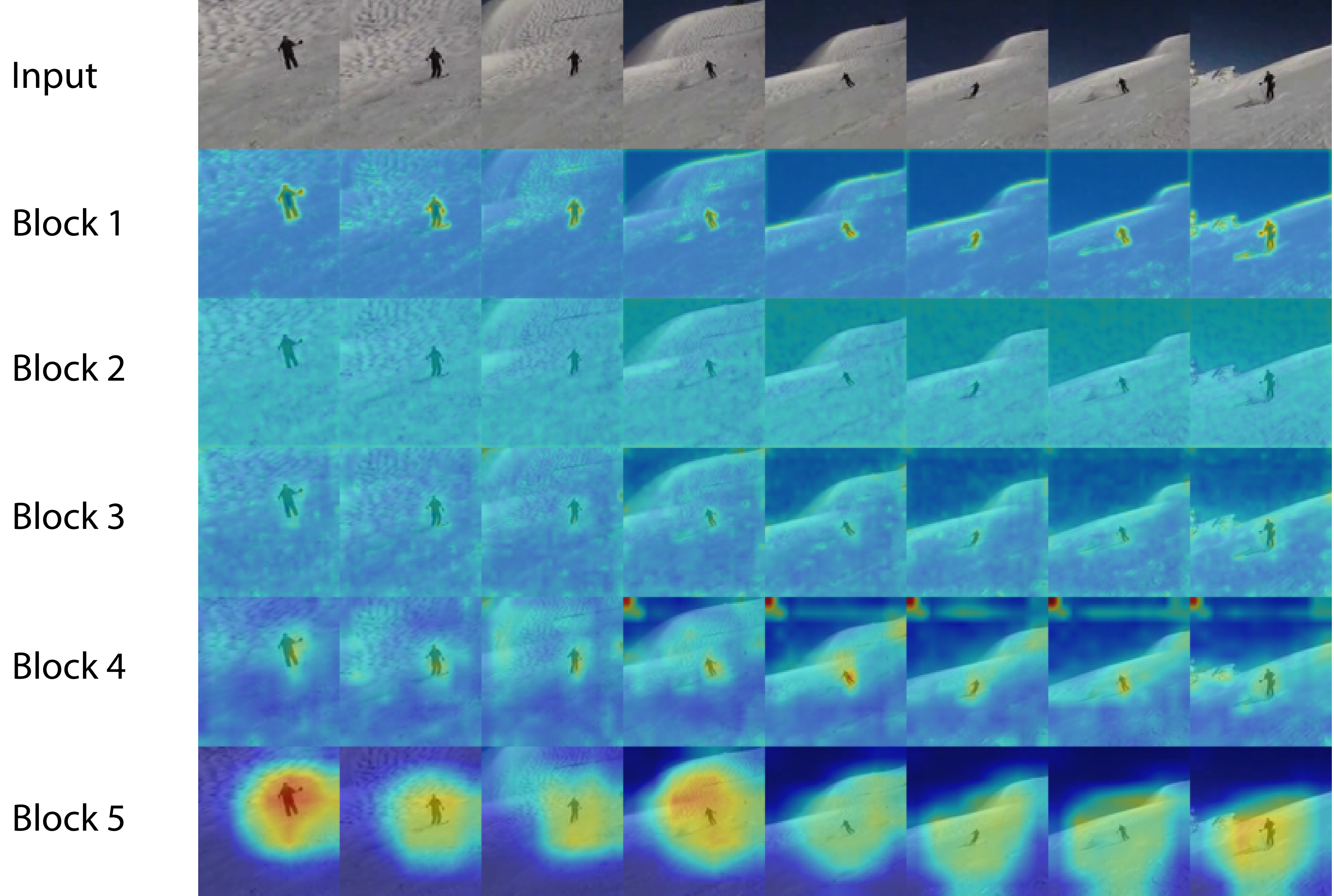}
		\caption{Visualization of the spatial attention in the "skiing" category. Red and blue represent greater and less attention, respectively. The first row is the original input, and the following rows are the attention results from each Resnet block.}
		\label{fig::SAT}
	\end{figure}

    The output of each intermediate layer should contain clues of the spatial relationships. The patterns captured by different features heavily vary in the output because of different focuses. However, there are some areas focused by a large number of features, which are the high-attention part we seek. What we need to do is to construct a mapping from the 3D output to a 2D attention image. The method of constructing the mapping from 3d output to 2d spatial attention is discussed in detail in \cite{zagoruyko2016paying}. We chose "Sum Function" because of its excellent performance, and it is very intuitive, as shown in Eq.\ref{eq1}. We define the following activation-based spatial attention map:
	
	\begin{equation}
	\label{eq1}
	F_{x,y}(O) = \sum_{i=1}^{C}|O_{x,y,i}|
	\end{equation}
	
	where $O$ means the 3D output of the network and $x$, $y$, $i$ represents the specific position of the output. The visualization of spatial attention is shown in Fig.~\ref{fig::SAT}.
	
	\subsubsection{Temporal Attention Definition}  
	
\begin{figure}
	    \includegraphics[width=0.48\textwidth]{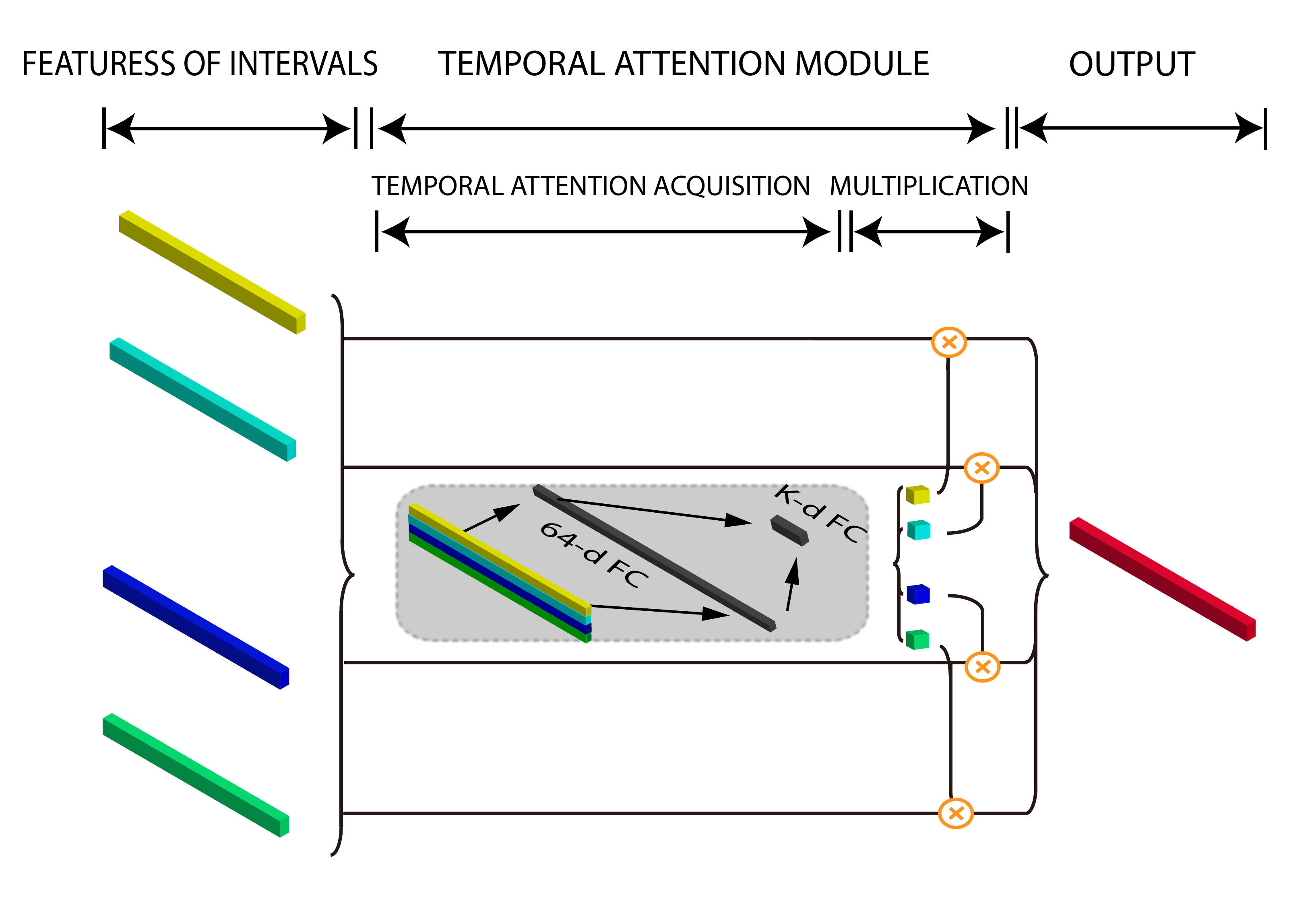}
	    \caption{Temporal Attention Module. Features of each interval are put together to generate their respective attention values, which are then multiplied by the original features to produce the final result.}
	    \label{fig::TAT}
    \end{figure}

	To capture the importance of each interval,  different from TSN~\cite{wang2016temporal}, we specially design a self-attention module, named temporal attention module (TAM), to estimate the importance of every interval, as shown in Fig.~\ref{fig::TAT}. 
	
	We divide a whole video into $K$ intervals and sample one or some consecutive frames from it as the input of the model. Interval is used as the basic unit. After the independent classification, we finally perform video-level prediction on all the results of $K$ intervals. 
	
	TAM takes the results of the feature extraction as input, processes data through two fully-connected (FC) layers, and finally normalizes the data results to [0,1] interval through a softmax layer. The output is assumed as the importance of each time interval, which is multiplied by the corresponding features to get the final video-level result.
	
	About the network detail of TAM, we use the feature of each interval as input and stack all features together to perform FC layers, which are 64-dimensional and K-dimensional, respectively. The former layer is used to expose more supervision signals and enhance the capability of analyzing data, while the latter would generate temporal attention.

	\subsubsection{Feature extraction}
	
	The backbone is needed to extract features from the input data. In our method, we use both a lighter 2D and a heavier 3D convolutional network. The 2D backbones always contains less parameters and need shorter inference time, such as VGG~\cite{simonyan2015very}, Resnet~\cite{he2016deep}, and Inception~\cite{inception}. In this case, we could analyze more intervals at the same time with limited memory.  Resnet-34 was used as the light 2D extractor, which is very fast and only has $3.6 * 10^{9}$ FLOPs. For the higher accuracy requirement, we also use a slow yet powerful model,  R2Plus1D-18. Its strong performance benefits from its novel structure replacing the 3D convolutional filters with a (2+1)D Resnet block.
	
	\subsection{Attention Transfer}
	
	The teacher-student paradigm is used to implement the transferring process, where the teacher network (T-Net) is fed with HR data, and the student (S-Net) is fed with eLR data. We also present the loss function of spatial-temporal attention transfer to capture the gap in attention between two networks.
	
	\subsubsection{Transfer Framework} 
	
	\begin{figure*}[t]
		\centering
		\includegraphics[width=1.0\linewidth]{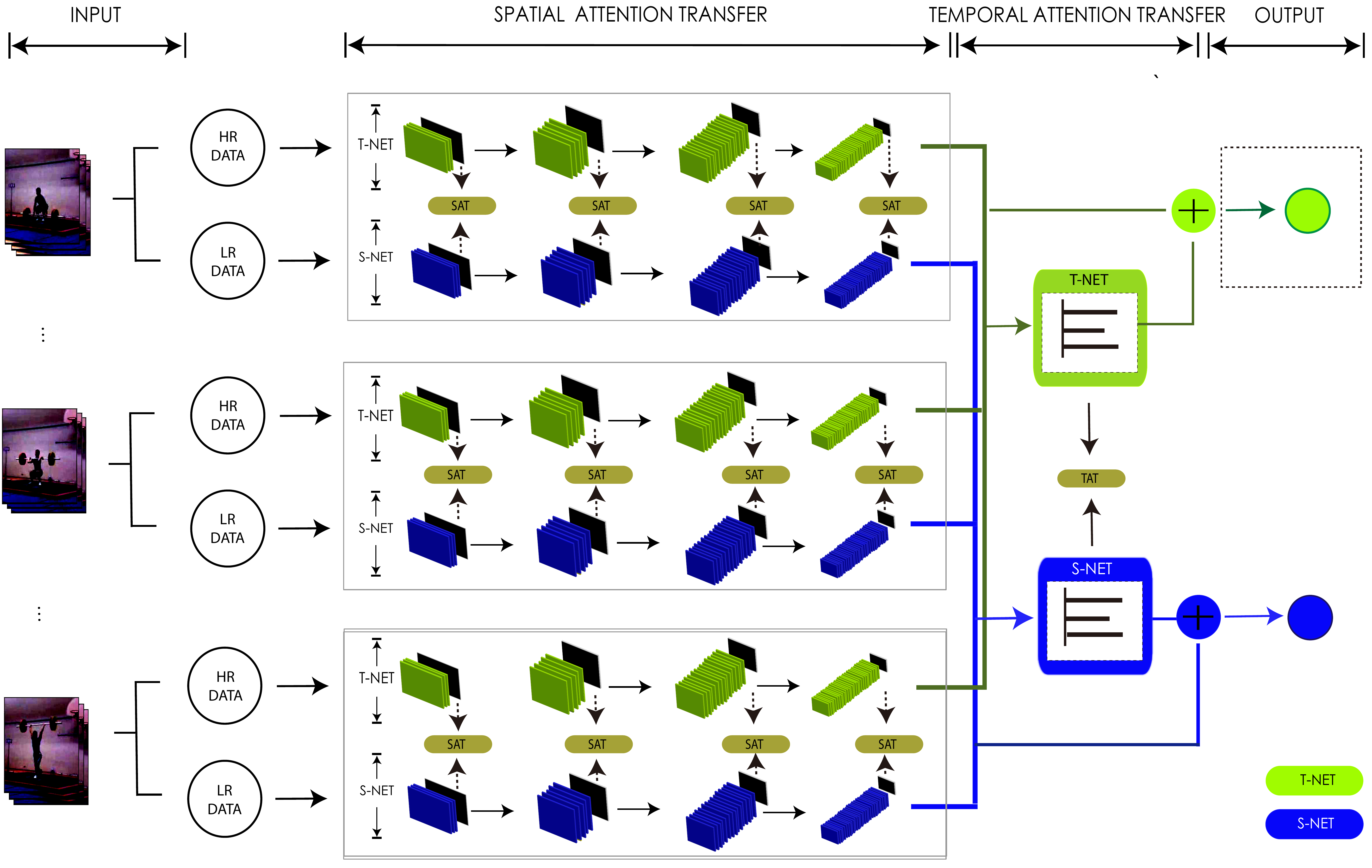}
		\caption{Overview of spatial-temporal attention transfer.  Given an $K$ intervals extracted from a video ($K$ is 3 in the figure), T-Net and S-Net generate spatial attention and time attention values simultaneously. The proposed STAT loss is used to force the attention difference to reduce.}
		\label{fig:network}
	\end{figure*}
	
	Two networks with the same structure but different-resolution data are used as T-Net and S-Net. More specifically, we input the HR data ($256 \times 340$) into the T-Net and the eLR data ($12\times16$) which are sampled from the HR data into S-Net. 
	Spatial-Temporal attention of two networks is obtained and can decrease the gap between them. With the decrease of attention gap, S-Net could better grasp what the network should pay attention to. The learning process is shown in Fig.~\ref{fig:network}.
	
	\subsubsection{Transfer Loss Function}
	
	The loss function of spatial-temporal attention transfer is composed of three parts, including the spatial transfer loss function, the temporal attention loss function, and a classification loss function.
	
	For computing spatial attention loss, we obtain both spatial attention from T-Net and S-Net firstly. Then, we minimize the output of the same index layer of the two networks to reduce the attention gap. T-Net with the more abundant data is more sensitive to the same action. In this case, the areas which the T-net focus on are more likely to be the key parts of action classification, and S-Net would obtain a higher performance if it notices the key parts better.
	
	However, the supervising signals from T-Net are very complex. If all weights are used to transfer, the training procedure is time-consuming. Besides, the T-Net does have certain parameters that deal with its own features, which may mislead the transferring. In this case, a selection for the transferring signal is necessary to accelerate the running speed. In our method, we only use the Resnet block output as the transferring signal. We used the cosine distance to describe the similarity of two features. Let S, T, and $ W^{S}$, $W^{T}$ denote student, teacher and their weights respectively. Let $Res$ denote the Resnet block output pairs of the two networks. Then the loss function of spatial attention transfer (SAT) is defined as the following:
	
	\begin{equation}
	L_{SAT}= \sum_{i \in Res }{\left\| 1 -  \frac{Q^{i}_{S} \cdot {Q^{i}_{T}}}{{\left\|Q^{i}_{S} \right\|}{\left\|Q^{i}_{T} \right\|}}\right\|_P}
	\end{equation}

	where $Q^{i}_{S}$ = $vec(F({A^{i}_{S}}))$ and $Q^{i}_{T}$ = $vec(F({A^{i}_{T}}))$ are the ${i}$-th pair of student and teacher attention maps in vectorized form, respectively, and $P$ refers to norm type. 
	
	For the temporary attention loss function, we focus on the discrepancy between the two TAM outputs, which is designed for estimating the importance of each time interval and importance the accuracy. With the module, we could concentrate on the period with more precise behavior information. In this loss function, the gap between two TAM outputs is the target to reduce, which could force S-Net to pay more attention to the high-informative interval. Except for the final FC layer, we also design a 64-dimensional hidden layer to magnify the differences in intermediate processing. As the spatial attention loss function, we also use cosine distance to measure the the similarity. The loss function of the temporal attention transfer (TAT) as following:
    
    \begin{equation}
	L_{TAT}= \sum_{j \in FC }{\left\| 1 - \frac{{Q^{j}_{S}}\cdot{Q^{j}_{T}}}{{\left\|Q^{j}_{S} \right\|}{\left\|Q^{j}_{T} \right\|}} \right\|_P}
	\end{equation}
	
	where $FC$ denotes the FC layer outputs pairs in temporal attention module.
	
	Finally, in addition to two loss functions mentioned above, we also add a loss function, cross-entropy, to help the network be sensitive to the final results. The cross-entropy loss (CE) is typically used for classification. Its value represents the gap between the final output and the label in video-level prediction. We define the total loss as:
	
	\begin{equation}
	L_{total}= W_{ce}\times L_{CE}  +  W_{sat}\times L_{SAT} + W_{tat}\times L_{TAT}  
	\end{equation}
	where $W_{ce}$, $W_{sat}$, $W_{tat}$  are the weight of the cross-entropy loss, the weight of spatial attention loss, and the weight of the temporal attention loss.
	
	\subsection{Confident Attention Transfer}
	
\begin{figure}[t]
		\includegraphics[width=1.0\linewidth]{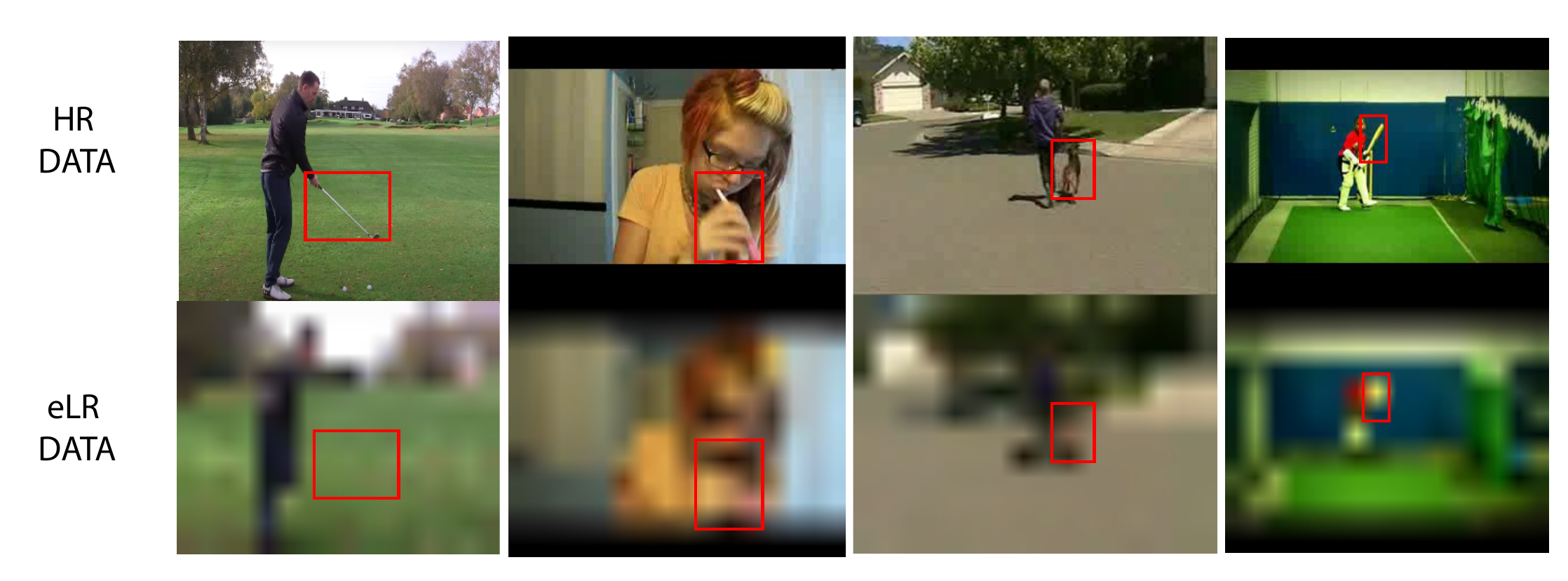}
		\caption{Visualization of the missing crucial information due to the down-sampling.}
		\label{fig:information loss}
	\end{figure}

	In the spatial-temporal attention transfer, the supervising signal from the teacher network is inaccurate sometimes. As Fig.~\ref{fig:information loss} shown, the red box in HR images would be blurred or even disappear in corresponding eLR images. In this situation, the supervision from HR data is meaningless or even misleading. For example, in the first column of Fig.~\ref{fig:information loss}, when the model determines whether it is a golfing action, the spatial space and the time interval where the player is swinging the golf club will provide more effective information. However, in the eLR feature space, the decisive parts cannot be perceptive, which causes a inaccurate supervised signal.
	
	In this part, we address this problem by introducing the confidence score. Specifically, we use the score as an indicator to measure the effectiveness of transfer processing. The high-informative areas and periods will be given a higher score, and the low-informative parts will be transferred with a lower score.
	
	\subsubsection{Confidence Score Maps}
	
	We consider three different methods to measure the confidence score, which is based on estimating the information loss degree of down-sampling. The first method is the difference-based, which was also used in \cite{wang2016temporal} to capture the changes over time. Then, we use a gradient-based method to obtain the degree of loss, which could make the high frequency change obviously. For the third approach, we use a multi-layer perception (MLP) as the extractor, which would be trained to seek the most suitable way to capture the loss. All methods would generate a binary map to indicate the information loss, and its inverse map is the confidence score map we need.
	
	
	\begin{figure}
		\centering
		
		\subfigure[Difference-based confidence score map]{
			\begin{minipage}[t]{1.0\linewidth}
				\centering
						\includegraphics[width=1.0\linewidth]{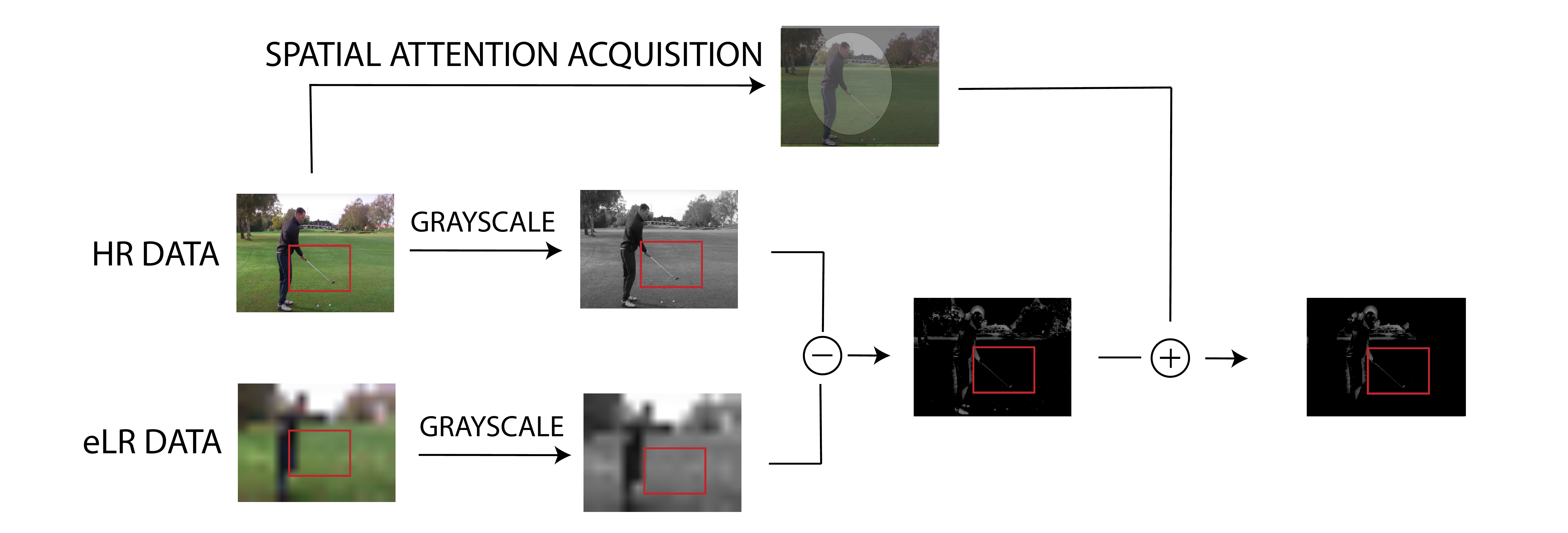}
			\end{minipage}%
		}%
	
		\subfigure[Gradient-based confidence score map]{
			\begin{minipage}[t]{1\linewidth}
				\centering		\includegraphics[width=1.0\linewidth]{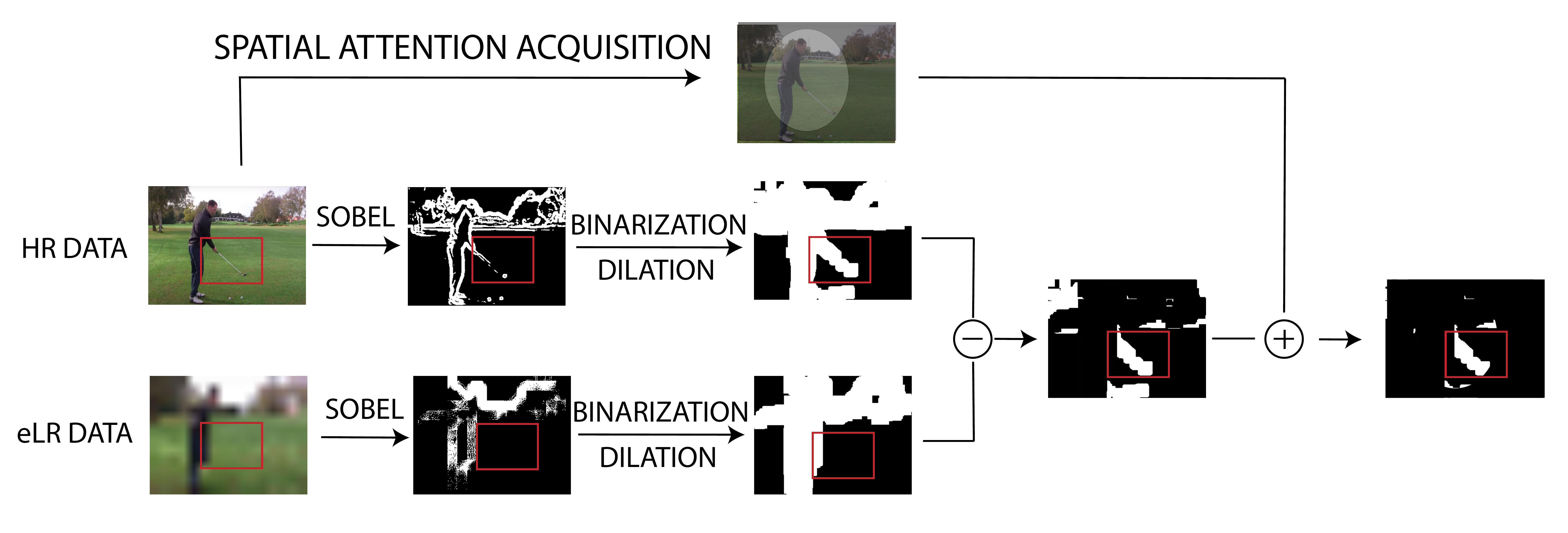}
			\end{minipage}%
		}%
	
		\subfigure[MLP-based confidence score map]{
			\begin{minipage}[t]{1\linewidth}
				\centering		\includegraphics[width=1.0\linewidth]{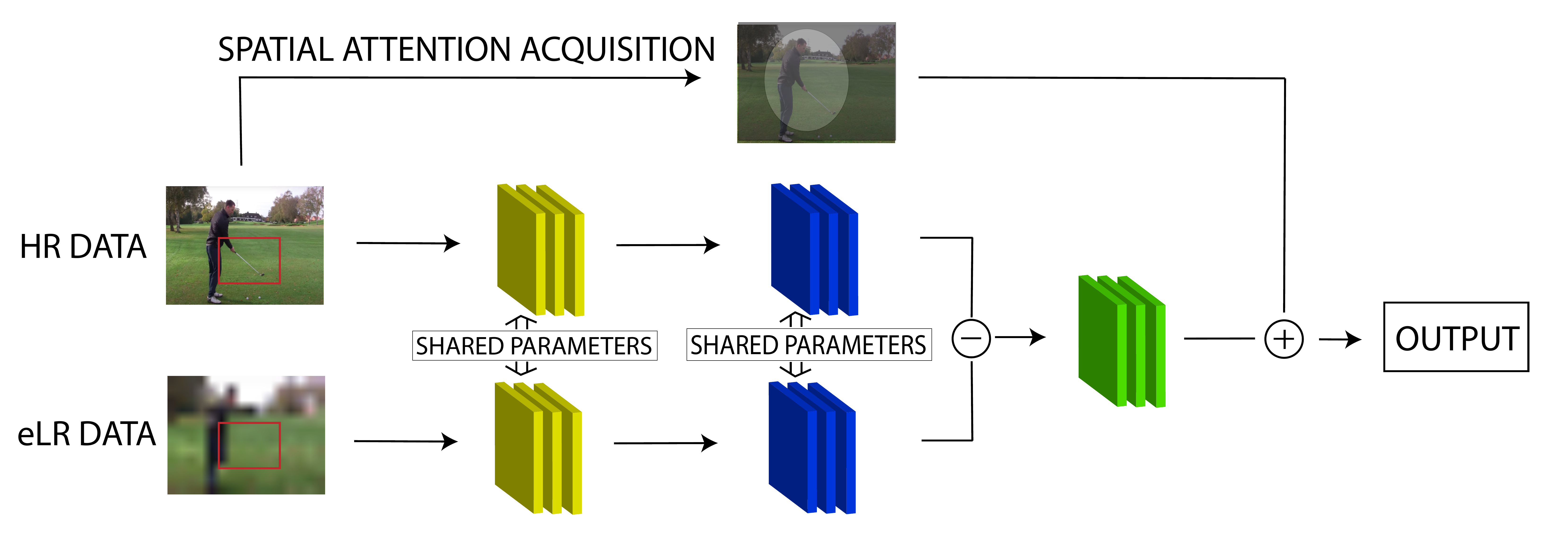}
			\end{minipage}
		}%
		
		\centering
		\caption{ Three methods to measuring the confidence of transferring signal.}
		\label{fig:measuring methods}
	\end{figure}
	
	\paragraph{Difference-based score map} As shown in the top of Fig.~\ref{fig:measuring methods}, the difference can effectively capture the diversity of two images. For example, \cite{wang2016temporal} used RGB-Diff to capture the difference between continuous frames, and the capturing method also could represent the changes between the HR and the eLR. In our proposed method, we use that as the reference for estimating the loss. The areas where the grey-scale value dramatically changed are worth estimating the information loss. In our practice, we directly subtract the image data of different resolutions and then normalize the generated difference output as its value of information loss.

	\paragraph{Gradient-based score map} As shown in the middle of  Fig.~\ref{fig:measuring methods}, the gradient can highlight the high-frequency parts and ignore the lower frequency areas. The gradient-based confidence map is more sensitive to changes in boundaries. In detail, we perform gradient extraction on image data of different resolutions and then enlarge the gradient result by dilation. Finally, we subtract two gradient maps and normalize them to obtain the corresponding information loss value.

	\paragraph{MLP-based score map} As shown in the bottom of  Fig.~\ref{fig:measuring methods}, the third approach is intended to find a novel way to learn the capturing information loss adaptively through a neural network. This method can seek a suitable feature extractor according to the feedback of the training result. For the MLP, we use two convolutional layers with K=3 filters as the extractor, and extractors of T-Net and S-Net share the parameters.
	
	\subsubsection{Confident Transferring Setting}
	
	To reduce the cost of calculation, the transferring only happens in high-information areas instead of the entire image. We set a threshold of the confidence score for the division of high-information areas. The areas where the confidence score below the threshold always means the original information dramatically changed and the supervision information provided is not reliable. The high-confidence areas represent the original information nearly unchanged, and the corresponding signal is more valuable. In our practice, the confidence score is used as the scalable factor which is multiplied with attention value in the transferring process. This can make the high-confidence signal have a higher credibility and reduce the interference of low-confidence transferring signal.
	
	The confident transferring of spatial attention is based on the involved areas and the value of transferring signal. Only areas with high information are involved in the transferring. The value of the transferring signal is obtained by multiplying the confidence score and the corresponding attention value. The spatial attention value will be obtained by mapping the 3D result of each layer output during the feature extraction process to a 2D plane. The improved method is more effective and faster in the process of transferring features. The confident spatial transferring loss function is almost the same as Equation (2) but the calculation range of Q, which are changed to $vec(F({A^{i} * C^{i}}))$, where C represents the corresponding confidence score of two i-pair networks.
	
	The confidence score in temporal attention transferring will be based on the corresponding spatial transferring process. The reason is that the damages caused by down-sampling only influence the quality of transferring spatial information, but still have a conductive impact in the temporal transferring. For all intervals in the same video, the objects (characters, props, etc.) would not change significantly, so all the share a confidence score. We use the ratio of the pixels without losing information in all the high-informative area as the temporal attention value, and the video-level score is the mean of all the intervals.
	
	\subsection{Training Strategies}
	
	In the training stage, there are two challenges. One is over-fitting due to the insufficient labeled data. Another problem is the contradiction between the importance of the supervisory signal and the training difficulty. Specifically, the deeper output is crucial for final prediction but hard to learn because that is based on the output of shallower layers.  
	
	\subsubsection{Unsupervised Training Method}
	\label{sec:3.4.1}
	
	Compared to single-frame pictures, video analysis application requires more data to conclude the temporal information. Moreover, the eLR model can not directly utilize existing pre-training weights because of its different low-level features from existing common datasets. In this case, a collection of more data seems to be a possible solution. However, manually labeling video is an extremely tedious task.
	
	\begin{figure}[h]
		\includegraphics[width=1.0\linewidth]{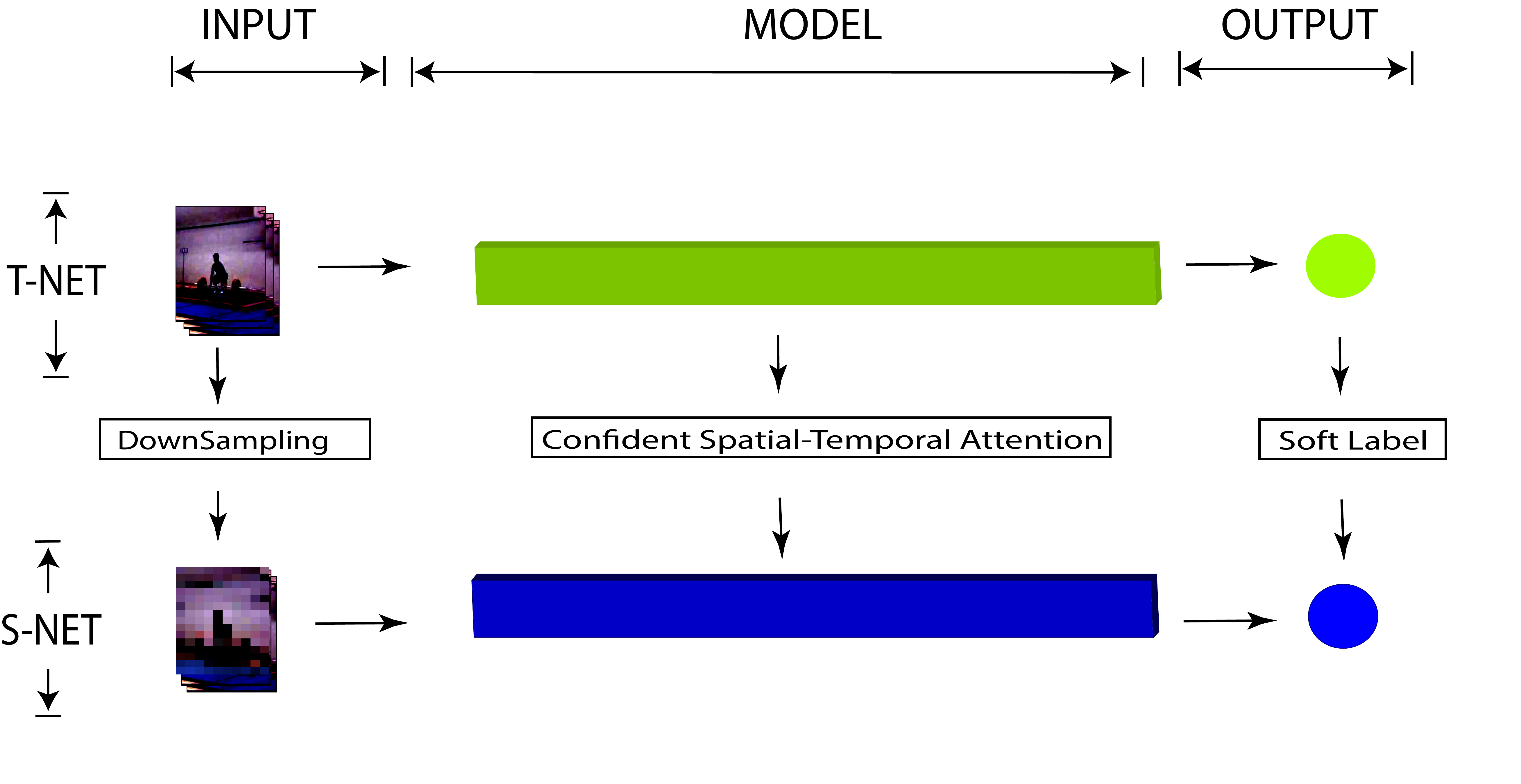}
		\caption{Overview of unsupervised training. The unlabeled are encoded to the spatial-attention information and the soft label by T-Net, and transferred to the S-Net.}
		\label{fig:unsupervised}
	\end{figure}

	As Fig.~\ref{fig:unsupervised} shown, we use an unsupervised approach to meet the enormous demand for data. That is using the T-Net with fixed parameters to encode data as the soft label.  There are specific requirements for training data; that is, the main body of action is human and contains a complete action rather than the cut-off videos. In our unsupervised training, we use the spatial-temporal information and soft label as the target of S-Net training.
	
	The unsupervised method could help the network overcome the problem. This is a more efficient way to use extra data, compared with the usage of pre-trained weights.
	
	\subsubsection{Rolling Weights}
	How to solve the contradiction between training difficulty and its importance?  We consider that the key to solve this problem is adopting different training strategies in different stages. The reason for the undesirable result of fixed weights is that, the feature differences in deeper layers has a large gap and are hard to learn. It is necessary to modify the weights of the shallower layer because deep information is based on the shallow part of the networks due to the different resolutions. The features that need to be extracted from the shallow layers of the two networks are not very close. To solve this problem, we propose a flexible and effective way to dramatically update the weights as following:
	
	\begin{equation}
	\label{eq5}
	W_{ij}= \frac{N - | i - j | }{\sum^{N}_{j=1} W_{ij}}
	\end{equation}
	
	where $N$ is the number of training stage. $i$, $j$ means the index of current block and training stage, respectively, $i, j\in N$. The strategy is adopted both in SAT and TAT.
	
	\section{Experiments}
	\label{sec:4}
	
	In this section, we evaluate and report the results of the proposed method of action recognition on eLR videos.
	
	\subsection{Datasets and Settings}

	We use the common datasets for the multi-class action classification, e.g. UCF101, HMDB51 and IXMAS. First down-sampling and then up-sampling is taken to obtain the eLR data.
	
	The HMDB51 dataset \cite{kuehne2011hmdb} is one of the most famous public datasets for action classification, which contains more than 7000 videos with 51 different action categories. The dataset is composed of the videos mostly collected from YouTube, including movie scenes. 
	
	The UCF101 dataset contains 101 \cite{soomro2012ucf101} classes and 13320 video clips. We follow the evaluation scheme of the THUMOS13 challenge and adopt the three train/test splits for evaluation. Each split in UCF101 includes about 9.5K training and 3.7K test videos.
	
	The IXMAS dataset \cite{weinland2010making} is comprised of 1,800 videos and 11 action classes that are incurred by occlusion, inter-class variation problem, and strong viewpoint changes.
	
    We resized these datasets to $12\times16$ using the average downsampling, while also include the lens blur term and the Gaussian noise term. For the videos with non-3:4 aspect ratio, a center cropping was used. 
	
	\subsection{Implementation Details}
	
	In our experiments, we used two backbones as the feature extraction module, Resnet-34 and R2Plus1D-18 for different reasons. In our practice, the input for the Resnet-34 is a single-frame data, and the whole video is divided into eight intervals. For R2Plus1D-18, the input from four intervals containing five consecutive frames each.
	
	Our entire training process is as follows: Firstly, a single model training is conducted, including T-Net with HR data and S-Net with eLR data. Then, a knowledge transfer model joint training is performed.
	
	In the single model training process, we use the mini-batch SGD to learn the network parameters, and the batch size is set to 256 and momentum is set to 0.9. The initial learning rate is 0.001, and the learning rate is divided by 10 for every 20 epochs. The first five epochs used the warm-up technique to find the direction of the gradient descent better. A total of 50 epochs were performed. Only cross-entropy was used as the loss function. We use the techniques of location jittering, horizontal flipping, cropping, and scale jittering for data augmentation. 
	
	Transferring learning is trained in the second stage. The norm $P$ in the transferring function is set to 2. We used different internal and external weights to control the role of each part in transferring, which will be discussed in the following part. We also use the unlabeled data to implement supervising learning, and we use the KINECTS without provided labels due to the reason described as Sec.~\ref{sec:3.4.1}. 
	
	There are some details in the confident transferring. In order to simplify the calculation, we regard the area where the difference is less than 24 as the same between the results of different resolutions. About the gradient-method, we use the Gaussian blur to ignore the unimportant parts selectively. We also used the dilation function from OpenCV with ten iterations. 
	
	The computational cost of the method proposed is as follows. The cost of lightweight mode is 6.51 GFLOPs and 21.34 M params, which takes a single frame as input. Another heavier model with 5-frame input needs more calculation resources, which is 28.99 GFLOPs and 33.43 M params.
	
	Note that, eLR data in our methods always means the upsampling eLR frames. That is, original videos are down-sampled to the required “extreme low resolution”, then used the bi-linear up-sampling method to restore the original size.  A larger input facilitates the subsequent operations, including subtraction, multiplication, convolution, etc. Even though the data are enlarged, the effective information is still limited.

	\subsection{Visualization of Spatial-temporal Attention}
	
		\begin{figure}[h]
		\includegraphics[width=1.0\linewidth]{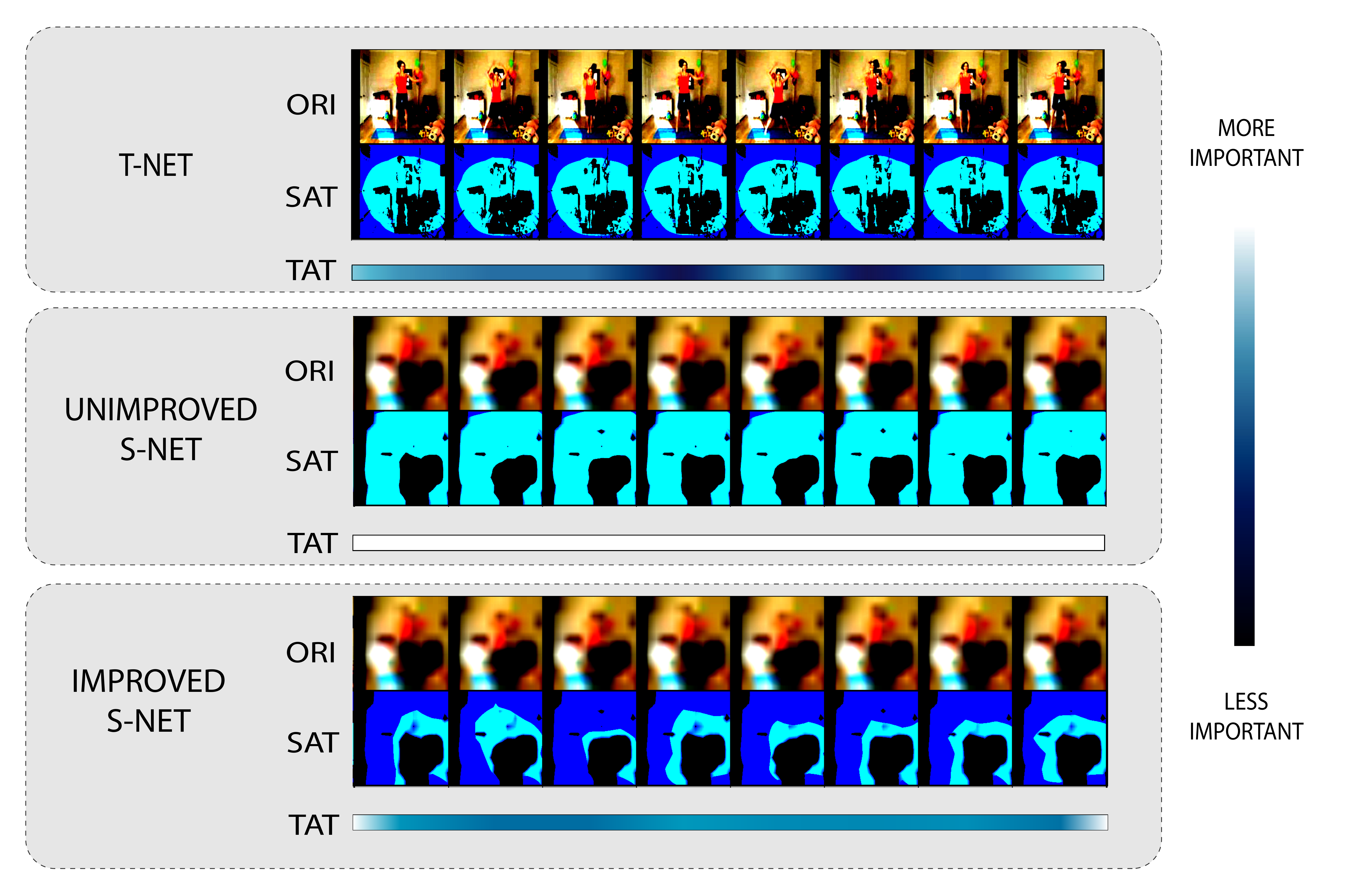}
		\caption{This figure consists of three parts. From top to bottom, T-Net results, unimproved S-Net results, and the improved S-Net are shown. The first row is the original image, the second row is the visualization of spatial attention of the last Resnet block, and the bar in the bottom is the visualization of temporal attention.}
		\label{fig::vis_attention}
	\end{figure}

	We visualize the spatial attention and temporal attention in this part. To have more intervals and make the effect of the temporal attention module more prominent, we use less-memory required Resnet-34 as the backbone. The UCF101 dataset is used as the training set and testing set. Only the RGB frames are used.
	
	In the spatial attention visualization, we randomly sample one frame from corresponding intervals as input. Then the input goes through four Resnet blocks. We perform the average operation on the output of each Resnet block in the channel dimension. Then the result is resized to the shape of the input by the bilinear interpolation, and we multiply it by the original image. In the temporal attention visualization, we put 512-dimensional features of each interval into the FC layers. Finally, the attention value for each interval is generated. We colored the temporal attention value in the bottom bar. 
	
	As shown in Fig.~\ref{fig::vis_attention}, we can determine the action is jumping. In the visualization of T-Net, the spatial attention is around the human, and the improved S-Net also obtains the information. However, the original algorithm focuses on the whole input. For temporal attention, the unimproved models can not capture the key intervals, and all the interval confidence are equal to one. In contrast, the improved model obtains a better result. In the fourth frame, the person is jumping with feet off the ground, and the interval obtains the highest confidence.
	
	\subsection{Visualization of Video Representation}
		\begin{figure}[b]
		\includegraphics[width=0.48\textwidth]{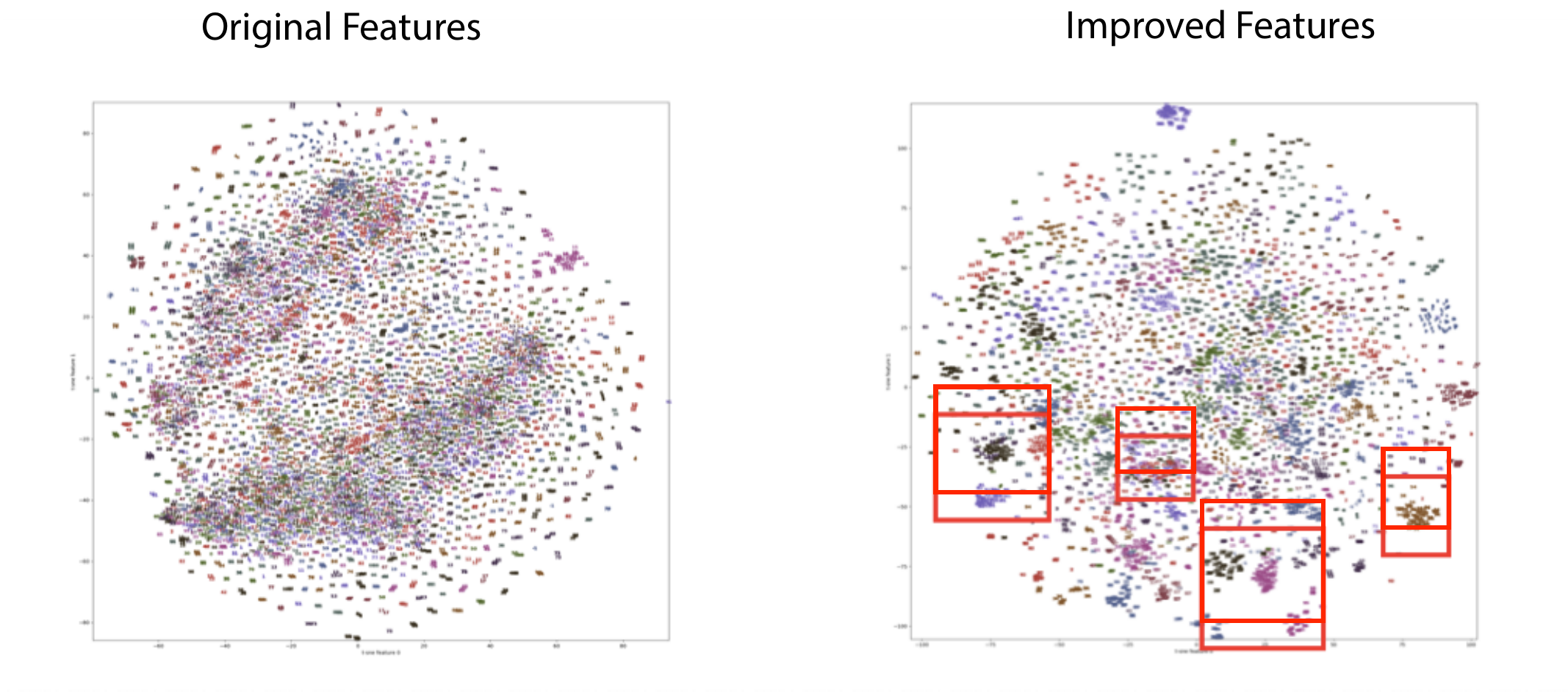}
		\caption{Video representation embedding visualizations on $12\times16$ UCF101 with Resnet-34. The left result is the original performance, while the right one is the improved one.}
		\label{fig::tsne}
	\end{figure}
	
	In this part, we visualize the representing result to show the improvement of our method. T-SNE technology is used to reduce dimension, which makes the visualization easy. Each item is visualized as one point and colors denote different actions. Fig.~\ref{fig::tsne} further shows our method aiming at focusing on the more informative part is better separated, even though not all the action classes perform well.

	\subsection{Comparison of Different Transferring Weights}	
	
	To seek the optimum weights setting and measure the influence of the weights, including the external weights and the internal weights, we conduct this experiment. In this part, the fixed strategy with different settings and rolling strategy both are tested.
	
	\begin{table}[h]
		\centering
		\caption{The results on $12 \times 16$ UCF101 dataset using Resnet-34 for different weights. The first four rows are fixed weight, and the last row is the rolling weight. The term External, In SAT, In TAT denote CE-SAT-TAT, Res1-Res2-Res3-Res4, FC1-FC2, respectively. }
		\begin{tabular}{c|c|c|c|c}
			\hline
			Type & External & In SAT & In TAT & Prec@1 \\
			\hline
			Fixed &	1/7 - 4/7 -2/7 & 1/4-1/4-1/4-1/4 &	1/2-1/2 & 46.12\% \\
			\hline
			Fixed &	4/7 - 1/7 -2/7 &	1/15-2/15-4/15-8/15 &	1/3-2/3 &	45.43\% \\
			\hline
			Fixed &	1/3 - 1/3 -1/3	& 1/4-1/4-1/4-1/4 &	1/2-1/2 &	\textbf{47.01\%} \\
			\hline
			Fixed &1/3 - 1/3 -1/3 &	1/15-2/15-4/15-8/15	& 1/3-2/3 &	46.60\% \\
			\hline
			Fixed & 1/3 - 1/3 -1/3 &	0-0-0-1 &	0-1	 &46.13\% \\
			\hline
			Rolling	& \multicolumn{3}{|c|}{As Eq.~\ref{eq5}}& \textbf{47.18\%} \\
			\hline
			
		\end{tabular}
		\label{tab::weight}
	\end{table}
	
	We conducted experiments on the above problems, and the experimental results are shown in Table ~\ref{tab::weight}.  We can see that, for the fixed weight, the best results are obtained when CE, SAT, and TAT are both 1/3. This can indicate that both SAT and TAT play supporting roles in the final result. For the weights in the middle of the SAT, the accuracy of the gradually-increase weight (1/15-2/15-4/15-8/15) is lower than the equal weight (1/4-1/4-1/4-1/4) of the weight at different overall weights, which shows that the weight setting according to the importance of the final result does not achieve the best. The reason is that the deep processing depends on the output of the shallow network. TAT weight setting is the same as the SAT. The rolling weight exceeds all the fixed weights because the contradiction between the importance of different weights and the dependence of adjacent weights is partly solved.
	
	\subsection{Comparison at Different Resolutions}
	
	Even though most of the relevant experiments are based on $12\times16$ resolution, we still believe that experiments at multiple resolutions are necessary for two reasons. Firstly, more resolution comparison experiments would provide valuable clues for balancing trade-off issues with resolution and accuracy in practice. Another reason is that we prove that our method is valid at all resolutions. We use UCF101 as the training and testing data and Resnet-34 as the extractor.
	
	\begin{table}[h]
		\centering
		\caption{The results on different resolution UCF101 dataset with our spatial-temporal attention transfer are shown.}
		
		\begin{tabular}{c|c|c|c|c}
			\hline
			~ Resolution &	 Before  &	 Before	 & After & After \\
			
			~ &	Prec@1 &	prec@5 &	Prec@1 & Prec@5 \\
			\hline
			($6\times8$) &	32.74 &	54.21 &	36.92 &	56.93 \\
			\hline
			($12\times16$)&43.13 &	68.17&46.19&68.29 \\
			\hline
			($18 \times 24$)&51.62&73.60&53.60&76.19 \\
			\hline
			($24 \times 32$)&55.08&79.52&56.30&80.22 \\
			\hline
			
		\end{tabular}
		\label{tab::resolution}
	\end{table}
	
	As shown in Table~\ref{tab::resolution}, our method works at different resolutions, and the improvement of the prec@1 gradually decreases with increasing resolution, 4.18\%, 3.06\%, 1.98\%, and 1.22\%, respectively. While the improvement of prec@5 is much smaller, from 2.72\% to 0.7\%. As the resolution increases, the accuracy of action recognition increases and the effect of TS learning decreases.
	
	\subsection{Comparison of Different Confidence Score Maps}
	
	In this part, we design an experiment to measure the performance of capturing information loss in three different methods. The amount of information loss can determine the credibility of the corresponding supervision signal, which is used as the weight of the signal during the transferring so that the supervision information provided by this method is more reliable. This process will directly affect the final result of the classification.
	
	The experimental setup is as follows: A 2d backbone and a 3d backbone are used to extract features for a shorter time granularity and a more powerful performance. The baseline method used is the basic spatial-temporal attention transfer and the attention acquisition part. Then we compare the difference-based, the gradient-based, and the MLP-based methods with the baseline, respectively. The dataset we use is 12x16 UCF101. Only RGB frames are used as the input because the TAM  enables the model to have the ability to analyze its time information.
	
	\begin{table}[h]
		\caption{The performance of different confidence score}
		\centering
		\begin{tabular}{c|c|c}
			\hline
			&	Resnet-34 & R2Plus1D-18 \\
			\hline
			Baseline  &	47.32\% &	52.72\% \\
			\hline
			Diffence-Based Method  & 47.71\% & 52.99\% \\
			\hline
			Gradient-Based Method  & 48.34\% &	53.51\% \\
			\hline
			MLP-Based Method  &	\textbf{49.53\%} &	\textbf{54.63\%} \\
			\hline
		\end{tabular}
		\label{tab::ILM}
	\end{table}
	
	As shown in Table~\ref{tab::ILM}, all three methods have a significant improvement over the baseline, which could strongly show the importance of confidence score and the effectiveness of our method. The difference-based method directly considers the gap with the different-resolution frames, and the accuracy compared with the baseline is limited,  increasing by 0.39\% and 0.27\%, respectively. The gradient-based method focus on high-frequency parts and the change of processing the original data brings a more noticeable improvement to the model, which has increased by 1.02\% and 0.79\%, respectively. The MLP-based method allows the network to adaptively adjust its way to process data through the supervision from the final classification. The improvement of the third method is the most obvious, increasing by 2.21\% and 1.91\%, respectively.
	
	\subsection{Exploration Study}
	
	In this subsection, we focus on our proposed learning method and its best practices. Two backbones are used, Resnet-34 and R2Plus1D-18. We did the relevant experiments to get the best practices. We used the 0.5-0.5-0, 0.5-0-0.5, 0.33-0.33-0.33 to represent SAT, TAT, SAT + TAT, respectively. Besides, we use the training strategy of rolling weight to make the weights easier to learn. Unsupervised learning is also added to the training because it avoids over-fitting.	
	
	\begin{table}[h]
		\centering
		\caption{The results with different parts of our method on the $12\times 16 $ UCF101 dataset. \textit{rc} and \textit{ut} mean the rolling weights and unsupervised training method, respectively. }
		
		\begin{tabular}{c|c|c}
			
			\hline
			&	Resnet-34&	R2Plus1D-18 \\
			\hline
			Teacher &	\textbf{78.82\%} &	\textbf{90.56\%} \\
			\hline
			Student &	42.79\% & 	48.95\% \\
			\hline
			SAT &	44.42\% &	52.03\% \\
			\hline
			TAT &	45.48\% &	50.90\% \\
			\hline
			STAT &	47.32\% &	52.72\% \\
			\hline
			CSTAT &	49.03\% &	54.46\% \\
			\hline
			CSTAT+\textit{rc} &	49.97\% &	54.78\% \\
			\hline
			CSTAT+\textit{rc} +\textit{ut} &	\textbf{50.78\%} &	\textbf{55.31\%} \\
			\hline
		\end{tabular}
		\label{tab::exploration}
	\end{table}
	
	As can be seen in Table~\ref{tab::exploration}, the performance is increased by 7.99\% and 6.36\%, respectively, after applying our proposed method. The latter is better than the former because T-Net has higher accuracy (78.82\% in Resnet-34, 90.56\% in R2Plus1D-18), which can provide more information. For the same reason, the latter network is also relatively stronger with eLR data than the former one. The accuracy of our model has been improved obviously on each network, from 42.79\% to 50.78\% on Resnet-34 and from 48.95\% to 55.31\% on R2Plus1D-18. It is worth noting that the TAT module has a smaller boost on R2Plus1D-18 than on Resnet-34. This is because of the different amount of intervals, and Resnet-34 can have a more precise range for sampling frames. The proposed rolling weight to solve the contradiction between the importance of different weights and the dependence of adjacent weights also helps with accuracy in both networks. Finally, unsupervised learning partially solved the problem of over-fitting by adding unlabeled data, which boosts the accuracy by 0.81\% in Resnet-34 and 0.53\% in R2Plus1D-18, respectively.
	
	\subsection{Comparison with the State-of-the-art}

	After assembling all the techniques described, we test it on the challenging datasets: $12\times16$ HMDB51. We used the experimental setting as \cite{Purwanto_2019_ICCV}, including 32-frame optical flow and RGB frames as input. R2Plus1D-34 is used as the feature extractor.

		\begin{table}[h]
		\caption{The performance of our model on the $12 \times 16$ HMDB51 dataset and IXMAS dataset.\\}
		\centering	
		\begin{tabular}{c|c|c}
			\hline
			Approach & HMDB51 Prec@1 & IXMAS Prec@1\\
			\hline
			PoT (CVPR'15)& 26.57 & 82.5 \% \\
			
			ISR (AAAI'17) & 28.68 & 85.1 \%\\
			
			Simonyan, et al  (WACV'17)&	 29.20 & 93.7 \%\\
			
			Ryoo,  et al  (AAAI'18)& 37.70 & 94.9 \%\\
			
			Xu, et al  (WACV'18) & 44.96 & 95.3 \%\\
			\hline
			I3D  (CVPR'17)  &52.61 & 93.89 \% \\
			Didik, et al (I3D) (ICCVW'19) & 57.84 &  97.22 \% \\
			\hline
			R2Plus1D-34  (CVPR'18)  &  52.29 & 94.2 \% \\		
			Ours(R2Plus1D-34) & \textbf{59.41} & \textbf{97.31} \%\\ 
			\hline
		\end{tabular}
		\label{tab::sota}
	\end{table}
	
	The results are summarized in Table~\ref{tab::sota}. It is not difficult to observe that our results outperform all state-of-the-art approaches on eLR action classification.

	\section{Discussion}
	
	Compared with other extreme low-resolution works, we have more clearly defined the recognition model needs from HR data, and our work achieved better results. The reason is that pre-trained weights trained on the HR data \cite{xu2018fully} or shared weight network with HR task \cite{chen2017semi} is not clear enough, which leads to a higher requirement for the complexity of the network and the amount of the input data. Besides,  the supervision provided by the HR model is leveraged more widely due to the whole-process transferring, rather than merely performing additional supervision from the beginning \cite{ryoo2018extreme} or the end. The unsupervised training method also exploits other data more effectively than other simple ways, like pre-training weights~\cite{chen2017semi}~\cite{xu2018fully}~\cite{ryoo2018extreme}, which makes our feature extraction better.
	
	Our work does have the areas for improvement in exploiting the available resources of HR data and training speed. Considerable resources for HR data are open source, but the eLR model cannot directly use that. For the pre-trained weights, there is a massive gap in the low-level features between HR data and eLR data, and the direct use of HR pre-trained weights without any processing would not bring the best results. So a numerical alteration or a well-designed network to change the HR pre-trained weights is worth furthermore researching. Besides, our running speed is slower because of the complicated network structure. How to maintain accuracy and reduce training time is worth exploring, such as pruning. 
	
	Confident spatial temporal attention transfer has achieved some results in the field of eLR motion recognition, which is 1.39\% higher than existing methods. This progress will make the motion recognition of far-view targets more accurate and enable the monitoring device to play a greater role. In addition, the success of eLR action recognition will also promote the popularization of extreme low-resolution recording devices to protect privacy.
	
	\section{Conclusion}
	
	In this paper, we explored how to make full use of the corresponding HD data to improve the accuracy of the eLR recognition model. We first designed a novel method for capturing spatial-temporal attention at different resolutions. After that, by forcing the eLR attention distribution close to the HR counterpart, the recognition model focuses on the more informative temporal interval and spatial part. Finally, the introduction of the confidence score has a selection for the supervision signal of the transferring process, thus ensuring the credibility of the signal. Our proposed method reaches the state-of-the-art level, which shows that the judgment clues in the HR data could reduce the difficulty in searching for effective parts in eLR data, and this information also plays a decisive role in the final classification.

	\newpage
	\bibliographystyle{IEEEtran}       
	\bibliography{bib}   
	
	%
	
	%
	
	\begin{IEEEbiographynophoto}{Yucai Bai} received the B.Eng. degree in software engineering from Chongqing University of Posts and Telecommunications, Chongqing, China, in 2017. He received the M.Eng. degree in software engineering with the college of Computer science, Sichuan University, Chengdu, China, in 2020. His research interests is transfer learning in computer vision, including action recognition, 3D semantic Map, image denosing. He has published a number of publications, which are accpted by IEEE Robotics and Automation Society (IROS), Nerurocomputing, etc.
	\end{IEEEbiographynophoto}

	\begin{IEEEbiographynophoto}{Qin Zou}
		(M’13–SM’19) received the B.E. degree in information science and the Ph.D. degree in computer vision from Wuhan University,
		Wuhan, China, in 2004 and 2012, respectively.
		From 2010 to 2011, he was a visiting Ph.D.
		student with the Computer Vision Lab, University of South Carolina, Columbia, SC, USA.
		He is currently an Associate Professor with the
		School of Computer Science, Wuhan University.
		His research activities involve computer vision,
		pattern recognition, and machine learning.
		Dr. Zou was a recipient of the National Technology Invention Award of
		China, in 2015. He is a member of the ACM.
	\end{IEEEbiographynophoto}

	\begin{IEEEbiographynophoto}{Xieyuanli Chen} is a Ph.D. student at the Photogrammetry and Robotics lab, University of Bonn, and also a member of the Technical Committee of RoboCup Rescue Robot League. He received his master's degree in Robotics in 2017 at the National University of Defense Technology, China, and his bachelor's degree in Electrical Engineering and Automation in 2015 at Hunan University, China. His research interests include localization, mapping and SLAM.
	\end{IEEEbiographynophoto}

	\begin{IEEEbiographynophoto}{Lingxi Li} (M’08-SM’13) received the B.Eng. degree in Automation from Tsinghua University, Beijing, China, in 2000, the M. S. degree in Control Theory and Control Engineering from the Institute of Automation, Chinese Academy of Sciences, Beijing, China, in 2003, and the Ph. D. degree in Electrical and Computer Engineering from the University of Illinois at Urbana-Champaign, in 2008. Since August 2008, he has been with Indiana University-Purdue University Indianapolis (IUPUI) where he is currently an Associate Professor of Electrical and Computer Engineering. Dr. Li’s current research focuses on the modeling, analysis, control, and optimization of complex systems; intelligent transportation systems; connected and automated vehicles; active safety systems; and human factors. He has authored/co-authored over one book and 100+ research articles in refereed journals and conferences and received two U.S. patents. Dr. Li received a number of awards including three conference best/remarkable paper awards, outstanding contributions award, outstanding editorial service award, and university research/teaching awards. He is currently serving as an associate editor for four IEEE Transactions/Journals and the Vice President of the IEEE Intelligent Transportation Systems Society. 

	\end{IEEEbiographynophoto}

	\begin{IEEEbiographynophoto}{Zhengming Ding} (S'14-M'18) received the B.Eng. degree in information security and the M.Eng. degree in computer software and theory from University of Electronic Science and Technology of China (UESTC), China, in 2010 and 2013, respectively. He received the Ph.D. degree from the Department of Electrical and Computer Engineering, Northeastern University, USA in 2018. He is a faculty member affiliated with Department of Computer Science, Tulane University since 2021. Prior that, he was a faculty member affiliated with Department of Computer, Information and Technology, Indiana University-Purdue University Indianapolis. His research interests include transfer learning, multi-view learning and deep learning. He received the National Institute of Justice Fellowship during 2016-2018. He was the recipients of the best paper award (SPIE 2016) and best paper candidate (ACM MM 2017). He is currently an Associate Editor of the Journal of Electronic Imaging (JEI) and IET Image Processing. He is a member of IEEE, ACM and AAAI.
	\end{IEEEbiographynophoto}

	\begin{IEEEbiographynophoto}{Long Chen}
		(M’13–SM’18) received the B.Sc. degree in communication engineering and the Ph.D. degree in signal and information processing from Wuhan University, Wuhan, China, in 2007 and 2013, respectively. From October 2010 to November 2012, he was co-trained Ph.D. Student with the National University of Singapore, Singapore. From 2008 to 2013, he was in charge of environmental perception system for autonomous vehicle SmartV-II with the Intelligent Vehicle Group, Wuhan University, Wuhan, China. He is currently an Associate Professor with the School of Data and Computer Science, Sun Yat-sen University, Guangzhou, China. His areas of interest includes perception system of intelligent vehicle.
	\end{IEEEbiographynophoto}

	

\end{document}